\pdfoutput=1

\documentclass[11pt]{article}

\usepackage{acl}

\usepackage{times}
\usepackage{latexsym}
\usepackage{multirow}
\usepackage{graphicx}
\usepackage{amsmath}
\usepackage{enumitem} 
\usepackage{mathtools}
\usepackage{amssymb}
\usepackage{xspace}
\usepackage{txfonts}
\usepackage{lipsum} 
\newcommand{\macrof}{$\mathrm{m}$-$\mathrm{F_1}$\xspace}
\newcommand{\microf}{$\mathrm{\muup}$-$\mathrm{F_1}$\xspace}
\newcommand{\mrp}{$\mathrm{m}$-$\mathrm{RP}$\xspace}

\newcommand{\cls}{\texttt{\small [cls]}\xspace}

\DeclareMathOperator{\sigmoid}{sigmoid}
\usepackage{xcolor,colortbl}
\definecolor{White}{HTML}{ffffff}
\definecolor{NiceGreen}{HTML}{34A853}
\definecolor{NiceRed}{HTML}{c83349}

\colorlet{c1}{White!93!NiceGreen}
\colorlet{c2}{White!91!NiceGreen}
\colorlet{c3}{White!89!NiceGreen}
\colorlet{c4}{White!87!NiceGreen}
\colorlet{c5}{White!85!NiceGreen}
\colorlet{c6}{White!83!NiceGreen}
\colorlet{c7}{White!81!NiceGreen}
\colorlet{c8}{White!79!NiceGreen}
\colorlet{c9}{White!77!NiceGreen}
\colorlet{c10}{White!75!NiceGreen}
\colorlet{c11}{White!73!NiceGreen}
\colorlet{c12}{White!71!NiceGreen}
\colorlet{c13}{White!69!NiceGreen}
\colorlet{c14}{White!67!NiceGreen}
\colorlet{c15}{White!65!NiceGreen}
\colorlet{c16}{White!63!NiceGreen}
\colorlet{c17}{White!61!NiceGreen}
\colorlet{c18}{White!59!NiceGreen}
\colorlet{c19}{White!57!NiceGreen}
\colorlet{c20}{White!55!NiceGreen}
\colorlet{c21}{White!53!NiceGreen}
\colorlet{c22}{White!51!NiceGreen}
\colorlet{c23}{White!49!NiceGreen}
\colorlet{c24}{White!47!NiceGreen}
\colorlet{c25}{White!45!NiceGreen}
\colorlet{c26}{White!43!NiceGreen}
\colorlet{c27}{White!41!NiceGreen}
\colorlet{c28}{White!39!NiceGreen}
\colorlet{c29}{White!37!NiceGreen}
\colorlet{c30}{White!35!NiceGreen}
\colorlet{c31}{White!33!NiceGreen}
\colorlet{c32}{White!31!NiceGreen}
\colorlet{c33}{White!29!NiceGreen}
\colorlet{c34}{White!27!NiceGreen}
\colorlet{c35}{White!25!NiceGreen}
\colorlet{c36}{White!23!NiceGreen}
\colorlet{c37}{White!21!NiceGreen}
\colorlet{c38}{White!19!NiceGreen}
\colorlet{c39}{White!17!NiceGreen}
\colorlet{c40}{White!15!NiceGreen}
\colorlet{c41}{White!13!NiceGreen}
\colorlet{c42}{White!11!NiceGreen}
\colorlet{c43}{White!9!NiceGreen}
\colorlet{c44}{White!7!NiceGreen}
\colorlet{c45}{White!5!NiceGreen}

\colorlet{n0}{White!100!NiceRed}
\colorlet{n1}{White!98!NiceRed}
\colorlet{n2}{White!96!NiceRed}
\colorlet{n3}{White!94!NiceRed}
\colorlet{n4}{White!92!NiceRed}
\colorlet{n5}{White!90!NiceRed}
\colorlet{n6}{White!88!NiceRed}
\colorlet{n7}{White!86!NiceRed}
\colorlet{n8}{White!84!NiceRed}
\colorlet{n9}{White!82!NiceRed}
\colorlet{n10}{White!80!NiceRed}
\colorlet{n11}{White!78!NiceRed}
\colorlet{n12}{White!76!NiceRed}
\colorlet{n13}{White!74!NiceRed}
\colorlet{n14}{White!72!NiceRed}
\colorlet{n15}{White!70!NiceRed}
\colorlet{n16}{White!68!NiceRed}
\colorlet{n17}{White!66!NiceRed}
\colorlet{n18}{White!64!NiceRed}
\colorlet{n19}{White!62!NiceRed}
\colorlet{n20}{White!60!NiceRed}
\colorlet{n21}{White!58!NiceRed}
\colorlet{n22}{White!56!NiceRed}
\colorlet{n23}{White!54!NiceRed}
\colorlet{n24}{White!52!NiceRed}
\colorlet{n25}{White!50!NiceRed}
\colorlet{n26}{White!48!NiceRed}
\colorlet{n27}{White!46!NiceRed}
\colorlet{n28}{White!44!NiceRed}
\colorlet{n29}{White!42!NiceRed}
\colorlet{n30}{White!40!NiceRed}
\colorlet{n31}{White!38!NiceRed}
\colorlet{n32}{White!36!NiceRed}
\colorlet{n33}{White!34!NiceRed}
\colorlet{n34}{White!32!NiceRed}
\colorlet{n35}{White!30!NiceRed}
\colorlet{n36}{White!28!NiceRed}
\colorlet{n37}{White!26!NiceRed}
\colorlet{n38}{White!24!NiceRed}
\colorlet{n39}{White!22!NiceRed}
\colorlet{n40}{White!20!NiceRed}
\colorlet{n41}{White!18!NiceRed}
\colorlet{n42}{White!16!NiceRed}
\colorlet{n43}{White!14!NiceRed}
\colorlet{n44}{White!12!NiceRed}
\colorlet{n45}{White!10!NiceRed}
\colorlet{n46}{White!8!NiceRed}
\colorlet{n47}{White!6!NiceRed}
\colorlet{n48}{White!4!NiceRed}
\colorlet{n49}{White!2!NiceRed}
\colorlet{n50}{White!0!NiceRed}

\usepackage[T1]{fontenc}

\usepackage[utf8]{inputenc}

\usepackage{microtype}

\title{Improved Multi-label Classification under Temporal Concept Drift: \\ Rethinking Group-Robust Algorithms in a Label-Wise Setting}

\author{Ilias Chalkidis and Anders Søgaard \\
  Department of Computer Science, University of Copenhagen, Denmark \\
  \texttt{[ilias.chalkidis,soegaard]@di.ku.dk} \\}

\begin{document}
\maketitle
\begin{abstract}
 In document classification for, e.g., legal and biomedical text, we often deal with hundreds of classes, including very infrequent ones, as well as temporal concept drift caused by the influence of real world events, e.g., policy changes, conflicts, or pandemics. 
Class imbalance and drift can sometimes be mitigated by resampling the training data to simulate (or compensate for) a known target distribution, but what if the target distribution is determined by unknown future events?
Instead of simply resampling uniformly to hedge our bets, we focus on the underlying optimization algorithms used to train such document classifiers and evaluate several group-robust optimization algorithms, initially proposed to mitigate group-level disparities. Reframing group-robust algorithms as adaptation algorithms under concept drift, we find that Invariant Risk Minimization and Spectral Decoupling outperform sampling-based approaches to class imbalance and concept drift, and lead to {\em much} better performance on minority classes. The effect is more pronounced the larger the label set. 
\end{abstract}

\section{Introduction}

Large-scale multi-label document classification is the task of assigning a subset of labels from a large predefined set -- of, say, hundreds or thousands of labels -- to a given document. Common applications include 
labeling scientific publications with concepts from ontologies \cite{tsatsaronis-etal-2015-bioasq}, associating medical records with diagnostic and procedure labels \cite{Johnson2017}, pairing legislation with relevant legal concepts \cite{Mencia2007}, or categorizing product descriptions \cite{Lewis2004}. The task in general presents interesting challenges due to the large label space and two-tiered skewed label distributions.

\begin{figure*}[t]
    \centering
    \includegraphics[width=\textwidth]{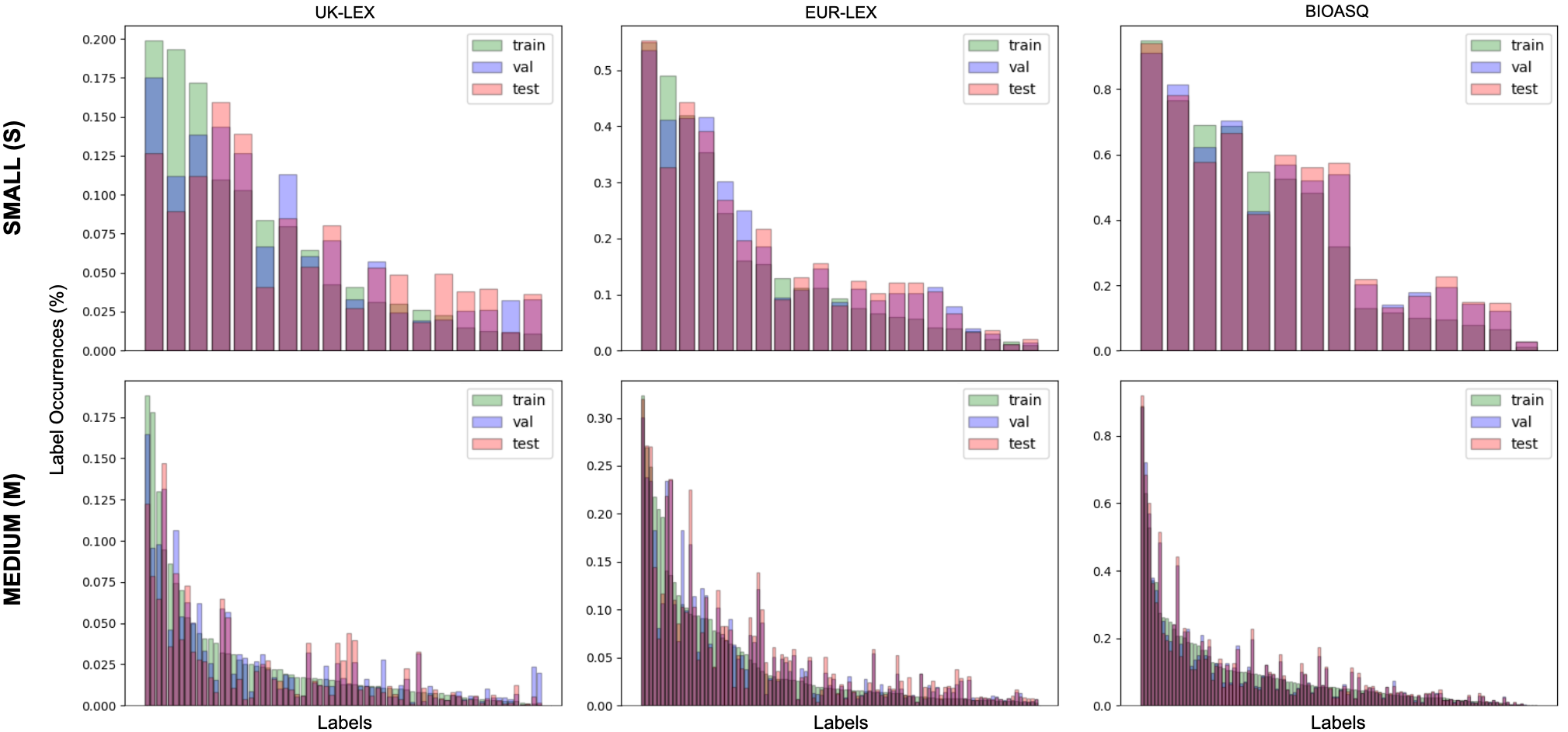}
    \vspace{-8mm}
    \caption{Label distributions of all datasets (UK-LEX, EUR-LEX, BIOASQ) and settings (small and medium sized label sets). Labels (bars) are ranked from most (left) to least (right) represented in the training set. \emph{Class imbalance} across labels in the $x$ axis and \emph{temporal concept drift} across subsets depicted with different coloured bars in the $y$ axis, i.e., a higher misalignment of color bars denotes a higher label distribution swift.}
    \label{fig:imbalance}
    \vspace{-4mm}
\end{figure*}

\paragraph{Class Imbalance} In multi-label classification, datasets often exhibit class imbalance, i.e., skewed label distributions (Figure~\ref{fig:imbalance}). Common methods include resampling and
reweighting based on heuristic assumptions, but methods are known to suffer from unstable performance, poor applicability, and high computational cost in complex tasks where their assumptions do not hold \cite{DBLP:conf/nips/0002W0C0020}. Datasets with long-tail frequency distributions, like the ones considered below -- sometimes referred to as {\em power-law datasets} \citep{rubin2012} -- can be particular challenging. Also, the heuristics fix the trade-off between exploiting as much of the training data as possible and balancing the classes, instead of trying to learn the optimal trade-off. 

\begin{figure}[t]
    \centering
    \includegraphics[width=\columnwidth]{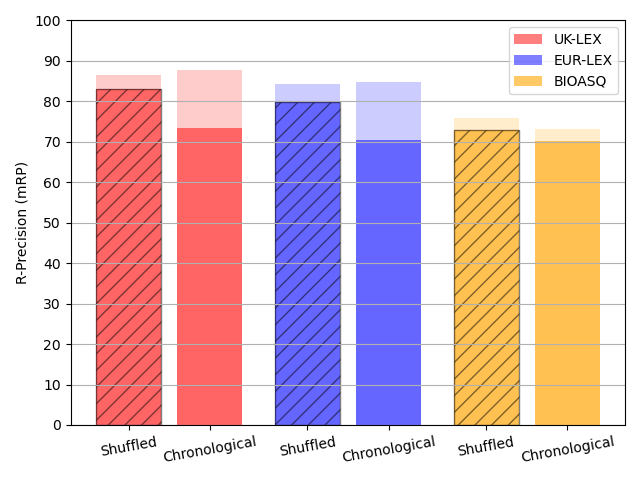}
    \vspace{-7mm}
    \caption{Model performance using \emph{random} vs. \emph{chronological} splits across the medium-sized datasets (Table~\ref{tab:datasets}). The shaded parts of the bars are the train/test discrepancy due to \emph{over-fitting}. The performance drop from random to chronological splits demonstrates the \emph{temporal concept drift}.}
    \label{fig:splits}
    \vspace{-5mm}
\end{figure}

\paragraph{Temporal Concept Drift} Moreover, class distributions may change over time. This is one dimension of the \emph{temporal generalization} problem \cite{lazaridou2021}. Recently, \citet{sogaard2021} argued chronological data splits are necessary to estimate real-world performance, contrary to random splits \cite{gorman-bedrick-2019-need}, because random splits artificially removes drift. Temporal concept drift, which we focus on here -- instead of covariate shift \cite{Shimodaira2000}, for example -- is an instance of concept drift \cite{10.1145/2523813}, often discussed in the domain adaptation literature, e.g., \citet{chan-ng-2006-estimating}.

\section{Related Work}

\paragraph{Temporal Drift}
Temporal drift has been studied in several NLP tasks, including document classification \cite{huang-paul-2018-examining,huang-paul-2019-neural-temporality}, sentiment analysis \cite{lukes-sogaard-2018-sentiment}, Named Entity Recognition (NER) \cite{rijhwani-preotiuc-pietro-2020-temporally}, Neural Machine Translation (NMT) \cite{levenberg-etal-2010-stream} and Language Modelling \cite{lazaridou2021}. None of these papers focus on class imbalance and temporal concept drift. These papers have mainly been diagnostic, not providing technical solutions that are applicable in our case.

\paragraph{Multi-label Class Imbalance}
Class imbalance in (large-scale) multi-label classification has so far been studied through the lens of network {\em architectures}, searching for the best neural architecture for handling few- and zero-shot labels in the multi-label setting. To improve the performance for underrepresented (few-shot) classes, \citet{snell-etal-2017} introduced Prototypical Networks that average all instances in each class to form {\em prototype} label vectors (encodings), a form of inductive bias, which improved few-shot learning. In a similar direction, \citet{Mullenbach2018} developed the Label-Wise Attention Network (LWAN) architecture, in which label-wise document representations are learned by attending to the most informative words for each label, using trainable label encodings (representations). 
\citet{rios-kavuluru-2018-shot} extended LWAN and the idea of {\em prototype} label encodings. They combined label descriptors with information from a graph convolutional network \cite{kipf2017semi} that considered the relations of the label hierarchy to improve the results in few-shot and zero-shot settings. Alternatives to LWAN were considered by \citet{chalkidis-etal-2020-empirical}, presenting minor improvements in the few-shot setting, but harming the overall performance.

\begin{table*}[t]
    \centering
    \resizebox{\textwidth}{!}{
    \begin{tabular}{l|l|c|lc|c|c}
        \bf Dataset & \bf Domain & \bf No. of Documents & \multicolumn{2}{c|}{\bf Setting} & \bf No. of Labels & \bf Distribution Swift (WS) \\ 
        \hline
         \multirow{2}{*}{UK-LEX (new)}& \multirow{2}{*}{UK Legislation} & \multirow{2}{*}{36,500} & Small & (S) & 18 / 18 & 8$\times$\\
         & & & Medium  & (M) & 69 / 69  & 5$\times$\\
         \hline
         \multirow{2}{*}{EUR-LEX \cite{chalkidis2021-multieurlex}} & \multirow{2}{*}{EU Legislation} & \multirow{2}{*}{65,000} & Small  & (S) & 20 / 21 & 9$\times$\\
         & & & Medium  & (M) & 100 / 127  & 7$\times$\\
         \hline
         \multirow{2}{*}{BIOASQ \cite{tsatsaronis-etal-2015-bioasq}} & \multirow{2}{*}{Biomedical Articles} & \multirow{2}{*}{100,000} & Small  & (S) & 16 / 16 & 29$\times$ \\
         & & & Medium  & (M) & 112 / 116  & 5$\times$\\
    \end{tabular}
    }
    \vspace{-2mm}
    \caption{Main characteristics of the examined datasets. We report the application domain, the number of documents, the available settings and the corresponding number of labels (used / total), and the label distribution swift between random and chronological splits using the Wasserstein Distance (WS) between train-test label probability distributions, i.e., $WS_{chronological} = N\times WS_{random}$.}
    \label{tab:datasets}
    \vspace{-3mm}
\end{table*}
\vspace{-2mm}
\paragraph{Robustness}
The literature on inducing robust models from skewed data is rapidly growing. See \citet{wilds2021} for a recent survey. The group-robust learning algorithms we adapt and evaluate, e.g., Group Distributionally Robust Optimization \cite{sagawa-etal-2020-dro}, are discussed in detail in Section~\ref{sec:algorithms}. 
Recent studies targeting fairness show that class imbalance has connections to bias \cite{blakeney2021measure,subramanian-etal-2021-fairness}, i.e., mitigating class-wise disparities has a chain effect on lowering group-wise disparities.

\paragraph{Main Contributions}  We focus on (large-scale) multi-label document classification and study a fundamental component of the learning process leading to performance disparities across labels, i.e., the underlying \emph{optimization algorithm} used for training. We consider group-robust optimization algorithms initially proposed to mitigate group disparities given specific attributes (e.g., gender, race), but re-frame these algorithms to optimize performance across labels rather than across groups.

\section{Datasets}
\label{sec:datasets}

We experiment with three datasets (Table~\ref{tab:datasets}) from two domains (legal and biomedical), which support two different classification settings (label granularities), i.e., label sets including more abstract or more specialized concepts (labels).\footnote{We originally also considered the MIMIC-III dataset of \citet{Johnson2017} including discharge summaries fro US hospitals annotated with ICD-9 medical codes, but the publication date of the documents has been ``counterfeited'' as part of the anonymization process. Experimental results with random splits are presented in Appendix~\ref{sec:mimic}.}

\paragraph{UK-LEX} United Kingdom (UK) legislation is publicly available as part of the United Kingdom's National Archives.\footnote{\url{https://www.legislation.gov.uk/}} Most of the laws have been categorized in thematic categories (e.g., health-care, finance, education, transportation, planing) that are presented in the document preamble and are used for archival indexing purposes. 

We release a new dataset, which comprises 36.5k UK laws (documents).\footnote{The UK-LEX dataset is available at \url{https://zenodo.org/record/6355465/}.} The dataset is chronologically split in training (20k, 1975--2002), development (8.5k, 2002--2008), test (8.5k, 2008--2018) subsets. We manually extract and cluster the topics to supports two different label granularities,  comprising 18, and 69 topics (labels), respectively.

\paragraph{EUR-LEX} European Union (EU) legislation is published in EUR-Lex.\footnote{\url{http://eur-lex.europa.eu/}} All EU laws are annotated by EU's Publications Office with multiple concepts from EuroVoc, a thesaurus maintained by the Publications Office.\footnote{\url{http://eurovoc.europa.eu/}} 
EuroVoc has been used to index documents in systems of EU institutions, e.g., in web legislative databases, such as EUR-Lex and CELLAR, the EU Publications Office’s common repository of metadata and content. 

We use the English part of the dataset of \citet{chalkidis2021-multieurlex}, which comprises 65k EU laws (documents).
\footnote{The EUR-LEX dataset is available at \url{https://hf.co/datasets/multi_eurlex}.} 
The dataset is chronologically split in training (55k, 1958--2010), development (5k, 2010--2012), test (5k, 2012--2016) subsets. It supports four different label granularities. 
We use the 1st and 2nd level of the EuroVoc taxonomy including 21 and 127 categories, respectively.

\paragraph{BIOASQ} The BIOASQ (Task A: Large-Scale Online Biomedical Semantic Indexing) dataset \cite{tsatsaronis-etal-2015-bioasq,bioasq2021} comprises biomedical articles from PubMed,\footnote{\url{https://pubmed.ncbi.nlm.nih.gov}} annotated with concepts from the Medical Subject Headings (MeSH) taxonomy.\footnote{\url{https://www.nlm.nih.gov/mesh/}} MeSH is a controlled and hierarchically-organized vocabulary produced by the National Library of Medicine. 
The current version of MeSH contains more than 29k concepts referring to various aspects of the biomedical research (e.g., Diseases, Chemicals and Drugs).
It is used for indexing, cataloging, and searching of biomedical and health-related information, e.g., in MEDLINE/PubMed, and the NLM databases. 

We use a subset of 100k documents derived from the latest version (v.2021) of the dataset.\footnote{The original BIOASQ dataset is available upon request at \url{http://participants-area.bioasq.org/datasets}.} 
We sub-sample documents in the period 2000-2021, and we consider chronologically split training (80k, 1964--2015), development (10k, 2015--2018), test (10k, 2018--2020) subsets. We use the 1st and 2nd levels of MeSH, including 16 and 116 categories.

\section{Fine-tuning Algorithms}
\label{sec:algorithms}

In our experiments, we rely on pre-trained English language models \cite{devlin-etal-2019-bert} and fine-tune these using different learning objectives. Our main goal during fine-tuning is to find a hypothesis ($h$) for which the risk $R(h)$ is minimal:\vspace{-2mm}
\begin{flalign}
h^{*} &= \arg\min_{{h\in{\mathcal{H}}}}R(h) \\
R(h) &= {\mathbf{E}}[\mathcal{L}(h(x),y)]
\end{flalign}

\noindent where $y$ are the targets (\emph{ground truth}) and $h(x) = \hat{y}$ is the system hypothesis (model's predictions).

Similar to previous studies, $R(h)$ is an expectation of the selected loss function ($\mathcal{L}$). In this work, we study multi-label text classification (Section~\ref{sec:datasets}), thus we aim to minimize the binary cross-entropy loss across $L$ classes:
\begin{equation}
   \mathcal{L}(x) =\ -y\log {\hat {y}}-(1-y)\log(1-{\hat {y}})
\end{equation}

\noindent\textbf{ERM}~\cite{vapnik-1992}, which stands for Empirical Risk Minimization, is the most standard and widely used optimization technique to train neural methods. The loss is calculated as follows:\vspace{-2mm}
\begin{equation}
    \mathcal{L}_{ERM} = \frac{1}{N} \sum_{i=1}^{N} \mathcal{L}(x_i)
\end{equation}

\noindent where $N$ is the number of instances (training examples) in a batch, and $\mathcal{L}_i$ is the loss per instance.\vspace{2mm}

Furthermore, we consider a representative selection of group-robust fine-tuning algorithms that try to mitigate performance disparities with respect to a given attribute ($A$), e.g., in a standard scenario that could be the gender of a document's author in sentiment analysis, or the background landscape in image classification. In our case, the attribute of interest is the labeling of the documents. The attribute is split into $G$ groups, which in our case are the classes ($G=L$). All algorithms rely on a balanced group sampler, i.e., an equal number($N_{g_i}$) of instances (samples) per group ($g_i$) are included at each batch. Most of the algorithms are built upon group-wise losses ($\mathcal{L}_{g_i}$), computed as follows:\vspace{-2mm}
\begin{equation}
    \mathcal{L}(g_i) = \frac{1}{N_{g_i}}\sum_{j=1}^{N_{g_i}} \mathcal{L}(x_j)
\end{equation}

In our case, contrary to previous applications of group-robust algorithms, the groups (classes) are not mutually exclusive (documents are tagged with multiple labels). Hence, the group sampler can only guarantee that \emph{at least} $N$ groups (labels) will be considered at each step, but most probably even more.
In this work, we examine the following group-robust algorithms in a label-wise fashion:\vspace{2mm}

\noindent\textbf{Group Uniform} is the more naive group robust algorithm that uses the average of the group-wise (label-wise) losses -all groups (labels) are considered equally important-, instead of the standard sample-wise average, as follows:\vspace{-2mm}

\begin{equation}
    \mathcal{L}_{GM} = \frac{1}{G}\sum_{i=1}^{G} \mathcal{L}(g_i)
    \label{eq:mean_loss}
\end{equation}

\noindent\textbf{Group DRO}~\cite{sagawa-etal-2020-dro}, stands for Group Distributionally Robust Optimization (DRO). Group DRO is an extension of the Group Uniform algorithm, where the group-wise (label-wise) losses are weighted inversely proportional to the group (label) performance. The total loss is calculated as follows:
\vspace{-2mm}
\begin{equation}
    \mathcal{L}_{DRO} = \sum_{i=1}^{G}w_{g_i} * \mathcal{L}(g_i)\textrm{, where}
    \label{eq:dro_loss}
\end{equation}\vspace{-5mm}
\begin{equation}
      w_{g_i} = \frac{1}{W}(\hat{w}_{g_i} * e^{L(g_i)}) \quad \textrm{and} \quad W = \sum_{i=1}^{G} w_{g_i}
    \label{eq:dro_loss_w}
\end{equation} 

\noindent where $G$ is the number of groups (labels), $\mathcal{L}_g$ are the averaged group-wise (label-wise) losses, $w_g$ are the group (label) weights, $\hat{w_g}$ are the group (label) weights as computed in the previous update step.\vspace{2mm}

\noindent\textbf{V-REx}~\cite{krueger-etal-2020-rex}, which stands for Risk Extrapolation, is yet another proposed group-robust optimization algorithm. \citet{krueger-etal-2020-rex} hypothesize that variation across training groups is representative of the variation later encountered at test time, so they also consider the variance across the group-wise (label-wise) losses. In V-REx the total loss is calculated as follows:\vspace{-2mm}
\begin{equation}
    \mathcal{L}_{REX} = \mathcal{L}_{ERM} + \lambda * \mathrm{Var}([\mathcal{L}_{g_1},\dots,\mathcal{L}_{g_G}])
\end{equation}

\noindent where $\mathrm{Var}$ is the variance among the group-wise (label-wise) losses, and $\lambda$, a weighting hyper-parameter scalar.\vspace{2mm}

\noindent\textbf{IRM}~\cite{arjovsky-etal-2020-irm}, which stands for Invariant Risk Minimization, mainly aims to penalize variance across multiple training dummy estimators across groups, i.e., performance cannot vary in samples that correspond to the same group. The total loss is computed as follows:\vspace{-2mm}
\begin{equation}
    \mathcal{L}_{IRM} = \frac{1}{G} \left(\sum_{i=1}^{G} \mathcal{L}(g_i) + \lambda * P(g_i)\right)
     \label{eq:irm_loss}
\end{equation}

\noindent where $\mathcal{L}_{gi}$ is the loss of the $i_{th}$ instance, which is part of the $g_{th}$ group (label). Refer to~\citet{arjovsky-etal-2020-irm} for a more detailed introduction of the group penalty terms ($P_{g}$).\vspace{2mm}

\noindent\textbf{Deep CORAL}~\cite{baochen-etal-2016-deepcoral}, minimizes the difference in second-order statistics (covariances) between the source and target feature activations. In practice, it introduces group-pair penalties:\vspace{-4mm}
\begin{equation}
    \mathcal{L}_{CORAL} = \mathcal{L}_{ERM} + \lambda * \frac{1}{G} \left(\sum_{i=1}^{G}P(g_i, g_{i+1})\right)
\end{equation}\vspace{-4mm}
\begin{equation}
    P(g_i, g_{i+1}) = [\overline{C_{g_i}} -\overline{C_{g_{i+1}}}]^2 + [\overline{X_{g_i}}-\overline{X_{g_{i+1}}}]^2
    \label{eq:coral_penalty}
\end{equation}

\noindent where $\overline{C_{g_i}}$ are the averaged covariances of the $i$th group and $\overline{X_{g_i}}$ are the averaged features (document representations) of the $i$th group, respectively. Refer to~\citet{baochen-etal-2016-deepcoral} for a more detailed introduction of the group penalty terms ($P_{g}$).\vspace{2mm}

\noindent\textbf{Spectral Decoupling}~\cite{pezeshki2020gradient} relies on the idea of \emph{Gradient Starvation}. \citeauthor{pezeshki2020gradient} state that a network could become over-confident in its predictions by capturing only one or a few dominant features. Thus, adding an L2 penalty on the network's logits ($\hat{y}_i$) provably decouples the fixed points of the dynamics. The total loss is computed as follows:\vspace{-2mm}
\begin{equation}
    \mathcal{L}_{SD} = \mathcal{L}_{ERM} + \lambda * \frac{1}{N}\sum_{i=1}^{N} {\hat{y}_i}^2
    \label{eq:sd_loss}
\end{equation}

\noindent In our work, we consider the aforementioned algorithms in a label-wise setting, instead of a group-wise setting given a protected attribute. In our case, $G=L$, where $L$ is the number of labels.

\begin{table*}[t]
    \centering
    \resizebox{\textwidth}{!}{
    \begin{tabular}{l|ccc|ccc|ccc|ccc|ccc|ccc}
        \multirow{3}{*}{\bf Algorithm} & \multicolumn{6}{c|}{\bf \textsc{UK-LEX}} & \multicolumn{6}{c|}{\bf \textsc{EUR-LEX}}  & \multicolumn{6}{c}{\bf \textsc{BIO-ASQ}}\\
         & \multicolumn{3}{c|}{ \emph{Small}} & \multicolumn{3}{c|}{ \emph{Medium}} & \multicolumn{3}{c|}{ \emph{Small}} & \multicolumn{3}{c|}{ \emph{Medium}} & \multicolumn{3}{c|}{ \emph{Small}} & \multicolumn{3}{c}{ \emph{Medium}} \\
        & \microf & \macrof & \mrp & \microf & \macrof & \mrp & \microf & \macrof & \mrp & \microf & \macrof & \mrp & \microf & \macrof & \mrp & \microf & \macrof & \mrp \\
        \hline
ERM             & \bf 80.4 &  75.3 &  83.6 &  66.7 &  36.6 &  73.2 &  79.1 &  64.7 &  \bf 83.9 &  68.1 &  40.7 &  71.7 &  \bf 86.0 &  74.9 &  \bf 87.6 &  68.6 &  47.1 &  70.4 \\    
ERM+GS          & 80.3 &  75.2 &  \cellcolor{c1} 84.1 &  69.4 &  \cellcolor{c2} 38.8 &  73.6 &  79.0 &  64.9 &  84.2 &  \cellcolor{c1} 69.3 &  \cellcolor{c13} 54.7 &  71.2 & 85.6 & 75.2 & \cellcolor{n1} 86.3 & 68.4 & \cellcolor{c2} 49.2 & \cellcolor{n1} 69.4 \\  
\hline
Group Uniform   & 79.7 &  75.3 &  84.5 &  \cellcolor{c2} 69.2 &  \cellcolor{c19} 56.1 &  \bf \cellcolor{c2} 75.7 &  78.6 &  \cellcolor{c3} 68.0 &  \cellcolor{n1} 82.4 &  68.9 &  \cellcolor{c9} 50.4 &  71.2 &  85.5 &  \cellcolor{c1} 76.5 &  87.0 &  68.9 &  \cellcolor{c5} 52.5 &  69.8 \\    
Group DRO       & \cellcolor{n1} 79.0 &  \cellcolor{n1} 73.5 &  84.3 &  \cellcolor{n5} 60.9 &  \cellcolor{n8} 28.5 &  \cellcolor{n3} 69.3 &  \cellcolor{n1} 77.9 &  65.7  & \cellcolor{n4} 79.6 &  \cellcolor{n4} 63.4 &  \cellcolor{n12} 27.8 &  \cellcolor{n8} 63.3 &  \cellcolor{n1} 84.4 &  \cellcolor{n1} 73.5 &  \cellcolor{n2} 85.0 &  \cellcolor{n20} 48.6 &  \cellcolor{n30} 16.9 &  \cellcolor{n21} 48.6 \\   
Deep CORAL      & 80.1 &  75.7 &  83.8 &  \cellcolor{c1} 68.2 &  \cellcolor{c4} 40.3 &  73.3 &  78.6 &  \cellcolor{c3} 68.0 &  \cellcolor{n1} 82.5 &  67.9 &  \cellcolor{c4} 45.2 &  \cellcolor{n1} 70.2 &  85.3 &  \cellcolor{c1} 75.4 &  86.2 &  \cellcolor{c1} 69.1 &  \cellcolor{c11} 56.1 &  70.1 \\    
V-REx             & 80.0 &  75.5 &  \cellcolor{c1} 84.6 &  \cellcolor{c1} 68.6 &  \cellcolor{c17} 53.7 &  \cellcolor{c1} 74.9 &  78.5 &  \cellcolor{c3} 67.9 &  \cellcolor{n1} 82.7 &  68.8 &  \cellcolor{c8} 49.2 &  \cellcolor{n2} 69.6 &  85.5 &  \cellcolor{c1} 76.6 &  87.1 &  68.6 &  \cellcolor{c2} 49.9 &  69.9 \\    
IRM             & \bf80.4 &  75.8 &  \cellcolor{c1} 84.7 &  \cellcolor{c2} 69.4 &  \cellcolor{c22} 59.6 &  \cellcolor{c2} 75.6 &  78.9 &  \cellcolor{c2} 67.6 &  83.2 &  \cellcolor{c2} 70.4 &  \bf \cellcolor{c14} 54.8 &  72.4 &  85.4 &  \cellcolor{c1} 76.4 &  86.9 &  \cellcolor{c1} 69.8 &  \cellcolor{c8} 55.9 &  70.5 \\    

SD              & 80.3 &  \bf \cellcolor{c1} 76.8 &  \bf \cellcolor{c1} 84.8 &  \bf \cellcolor{c3} 70.0 &  \bf \cellcolor{c23} 59.8 &  \cellcolor{c2} 75.2 &  \bf 79.3 &  \bf \bf \cellcolor{c4} 68.9 &  \cellcolor{n4} 79.4 & \bf \cellcolor{c2} 70.8 &  \cellcolor{c11} 52.5 &  \bf \cellcolor{c1} 72.7 &  85.6 &  \bf \cellcolor{c2} 77.2 &  86.9 &  \bf \cellcolor{c2} 71.1 &  \bf \cellcolor{c6} 53.8 &  \bf \cellcolor{c1} 72.3 \\    

    \end{tabular}
    }
    \vspace{-2mm}
    \caption{\textbf{Overall} test results  of the \textbf{group-robust (label-robust) algorithms} across all datasets (UK-LEX, EUR-LEX, BIOASQ) and settings (small and medium sized label sets).}
    \label{tab:overall}
    \vspace{-4mm}
\end{table*}

\section{Experimental SetUp}

\paragraph{Baseline Models} For both legal datasets (UK-LEX, EUR-LEX), we use the small LEGAL-BERT model of \citet{chalkidis-etal-2020-legalbert}, a BERT \cite{devlin-etal-2019-bert} model pre-trained on English legal corpora. For BIOASQ, we use the small English BERT model of \citet{turc-etal-2019}. Following \citet{devlin-etal-2019-bert}, we feed each document to the pre-trained model and obtain the top-level representation $h_{\text{\cls}}$ of the special \cls token as the document representation. The latter goes through a dense layer of $L$ output units, one per label, followed by a sigmoid activation.

We also experiment with the Label-Wise Attention Network (LWAN) relying on a BERT encoder \cite{chalkidis-etal-2020-empirical}, dubbed BERT-LWAN.\footnote{The original model was proposed by \citet{Mullenbach2018}, with a CNN encoder.} \citeauthor{chalkidis-etal-2020-empirical} reported state-of-art results in EUR-LEX and AMAZON-13K using BERT-LWAN compared to several baselines. BERT-LWAN uses one  attention head per label to generate $L$ document representations $d_l$:
\vspace{-1mm}
\begin{flalign}
a_{lt} &= \frac{\mathrm{exp}(K(h_t) Q_{l})}{\sum_{t'} \mathrm{exp}(K(h_{t'}) Q_{l})} \\
d_l &=  \frac{1}{T} \sum^T_{t=1} a_{lt} V(h_t)
\end{flalign}

\noindent $T$ is the document length in tokens, $h_t$ the context-aware representation of the $t$-th token, $K$, $V$ are linear transformations of $h_t$, and $Q_l$ a trainable vector used to compute the attention scores of the $l$-th attention head; $Q_l$ can also be viewed as a label representation. Intuitively, each head focuses on possibly different tokens of the document to decide if the corresponding label should be assigned. BERT-LWAN employs $L$ linear layers ($o_l$) with $\sigmoid$ activations, each operating on a different label-wise document representation $d_l$, to produce the probability of the corresponding label $p_l$:
\vspace{-1mm}
\begin{flalign}
p_l = \sigmoid(d_l \cdot o_l)
\label{eq:lwan_outputs}
\end{flalign}

Across experiments, we use BERT models following a small configuration (6 transformer blocks, 512 hidden units and 8 attention heads), which allows us to increase the batch size up to 64 and consider samples with multiple labels (groups) in the group robust algorithms. In practice, this enables us to sample at least 4 samples per group (label) for all labels in the small label sets, and at least 1 sample per group (label) for 64 labels in the medium-sized label sets (69-112 labels).

\paragraph{Training Details}  We fine-tune all models using the AdamW \cite{loshchilov2018decoupled} optimizer with a learning rate of 2e-5. We use a batch size of 64 and train models for up to 20 epochs using early stopping on the development set.  We run three repetitions with different random seeds and report the test scores based on the seed with the best scores on development data. We report development scores on Appendix~\ref{sec:dev_results}.

\paragraph{Evaluation Metrics}
Given the large number and skewed distribution of labels, retrieval measures have been favored in large-scale multi-label text classification literature \cite{Mullenbach2018, You2019, chalkidis-etal-2020-empirical}. Following \citet{chalkidis-etal-2020-empirical}, we report \emph{mean R-Precision} (\mrp) \cite{manning2009}, while we also report the standard \emph{micro-F1} (\microf) and \emph{macro-F1} (\macrof) to better estimate the class-wise performance disparity.

\paragraph{Data and Code}

In our experiments, we extend the WILDs \cite{wilds2021} library, which provides an experimental framework for experimenting with group-robust algorithms. We effectively rewrote all parts of code to consider label-wise groups and losses, while we also implemented the unsupported methods (Group Uniform, V-REx, and Spectral Decoupling). For reproducibility and further exploration with new group-robust methods, we release our code on Github.\footnote{\url{https://github.com/coastalcph/lw-robust}} 

\begin{table}[t]
    \centering
    \resizebox{\columnwidth}{!}{
    \begin{tabular}{l|ccc|ccc}
        \multirow{2}{*}{Dataset} & \multicolumn{3}{c|}{Random} & \multicolumn{3}{c}{Chronological} \\
           & \microf & \macrof & \mrp & \microf & \macrof & \mrp \\
         UK-LEX SM) & \cellcolor{c6} \bf 89.3 & \cellcolor{c12} \bf 87.5 & \cellcolor{c10} \bf 92.9 & 80.4 & 75.3 & 83.6 \\
         UK-LEX (M) & \cellcolor{c8} \bf 78.2 & \cellcolor{c10} \bf 45.6 & \cellcolor{c12}\bf 85.0 & 69.2 & 36.6 & 73.2 \\
         \hline
         EUR-LEX (S) & \cellcolor{c8} \bf 86.8 & \cellcolor{c12} \bf 76.5 & \cellcolor{c6} \bf 89.5 & 79.3 & 64.4 & 84.2 \\
         EUR-LEX (M) & \cellcolor{c12} \bf 77.6 & \cellcolor{c10} \bf 49.8 & \cellcolor{c10} \bf 79.8 & 68.4 & 40.4 & 70.5 \\
         \hline
         BIOASQ (S) & \bf 86.5 & \bf 75.9 & \bf 88.8 & 86.0 & 74.9 & 87.6 \\
         BIOASQ (M) & \cellcolor{c2} \bf 71.9 & \cellcolor{c2} \bf 48.2 & \cellcolor{c2} \bf 72.3 & 68.6 & 47.1 & 70.4 \\
    \end{tabular}
    }
    \caption{Test results across all datasets and settings using \textbf{random vs. chronological splits} with ERM.}
    \label{tab:data_splits}
    \vspace{-4mm}
\end{table}

\section{Results}
\label{sec:results}

\begin{table*}[t]
    \centering
    \resizebox{\textwidth}{!}{
    \begin{tabular}{l|ccc|ccc|ccc|ccc|ccc|ccc}
        \multirow{2}{*}{\bf Algorithm}  & \multicolumn{6}{c|}{\bf \textsc{UK-LEX}} & \multicolumn{6}{c|}{\bf \textsc{EUR-LEX}} & \multicolumn{6}{c}{\bf  \textsc{BIOASQ}} \\
        & \multicolumn{3}{c|}{\emph{Head}} & \multicolumn{3}{c|}{\emph{Tail}} & \multicolumn{3}{c|}{\emph{Head}} & \multicolumn{3}{c|}{\emph{Tail}} & \multicolumn{3}{c|}{\emph{Head}} & \multicolumn{3}{c}{\emph{Tail}} \\
        & \microf & \macrof & \mrp & \microf & \macrof & \mrp & \microf & \macrof & \mrp & \microf & \macrof & \mrp & \microf & \macrof & \mrp & \microf & \macrof & \mrp \\
        \hline
ERM             & 71.8 &  55.7 &  77.2 &  38.4 &  17.0 &  76.2 &  73.4 &  61.9 &  75.7 &  27.5 &  19.4 &  51.7 &  71.7 &  60.6 &  73.3 &  46.2 &  33.6 &  58.2 \\ 
ERM+GS          & \bf \cellcolor{c1} 72.7 &  \cellcolor{c3} 58.4 &  \cellcolor{c1} 77.6 &  \cellcolor{c4} 42.6 &  \cellcolor{c13} 29.8 &  \cellcolor{c2} 77.7 &  73.3 &  \cellcolor{c1} 63.9 &  \cellcolor{n1} 74.2 &  \cellcolor{c20} 48.1 &  \cellcolor{c26} 45.4 &  \cellcolor{c4} 56.3 & \cellcolor{c1} 72.3 & \cellcolor{c1} 61.2 & \cellcolor{n1} 72.8 & \cellcolor{c2} 48.1 & \cellcolor{c7} 40.2 & 57.9 \\ 
\hline
Group Uniform   & 71.2 &  \cellcolor{c4} 60.4 &  \cellcolor{c1} 78.5 &  \cellcolor{c23} 62.1 &  \cellcolor{c34} 51.7 &  \cellcolor{c3} 79.8 &  73.4 &  62.3 &  74.7 &  \cellcolor{c15} 42.7 &  \cellcolor{c19} 38.5 &  \cellcolor{c1} 53.0 &  71.7 &  61.0 &  72.8 &  \cellcolor{c4} 51.1 &  \cellcolor{c10} 44.0 &  57.7 \\ 
Group DRO       & \cellcolor{n4} 66.9 &  \cellcolor{n9} 45.9 &  \cellcolor{n3} 73.6 &  \cellcolor{n9} 28.9 &  \cellcolor{n6} 10.6 &  \cellcolor{n6} 69.3 &  \cellcolor{n3} 70.0 &  \cellcolor{n11} 50.6 &  \cellcolor{n5} 70.2 &  \cellcolor{n20} 7.1 &  \cellcolor{n14} 4.9 &  \cellcolor{n22} 28.8 &  \cellcolor{n7} 63.9 &  \cellcolor{n27} 33.0 &  \cellcolor{n7} 65.5 &  \cellcolor{n45} 0.9 &  \cellcolor{n32} 0.7 &  \cellcolor{n50} 0.7 \\ 
Deep CORAL      & 69.2 &  \cellcolor{c6} 61.3 &  \cellcolor{n1} 76.5 & \cellcolor{c23} 62.0 &  \cellcolor{c11} 48.4 &  \cellcolor{c4} 80.0 &  72.6 &  \cellcolor{n1} 60.1 &  \cellcolor{n2} 73.4 &  \cellcolor{c8} 35.8 &  \cellcolor{c10} 30.4 &  \cellcolor{c5} 56.8 &  \cellcolor{c1} 72.7 & \cellcolor{c2} 63.1 & 73.7 & \cellcolor{c7} 52.3 & \cellcolor{c13} 46.5 & \cellcolor{c1} 59.2 \\
V-REx             & \cellcolor{n1} 70.2 &  56.6 &  76.9 &  \cellcolor{c23} 62.1 &  \cellcolor{c33} 50.7 &  \cellcolor{c5} 82.0 &  73.1 &  61.7 &  \cellcolor{n2} 73.3 &  \cellcolor{c15} 42.6 &  \cellcolor{c17} 36.8 &  \cellcolor{c3} 55.3 &  71.5 &  60.3 &  72.7 &  \cellcolor{c2} 48.8 &  \cellcolor{c5} 39.4 &  57.8 \\ 
IRM             & 71.4 &  \bf \cellcolor{c7} 62.8 &  \bf \cellcolor{c1} 78.6 &  \cellcolor{c23} 62.2 &  \cellcolor{c39} 56.3 &  \cellcolor{c4} 80.3 &  74.4 &  \bf \cellcolor{c2} 64.4 &  75.2 & \bf  \cellcolor{c21} 48.7 & \bf  \cellcolor{c25} 45.1 &  \cellcolor{c4} 56.5 &  72.3 &  \cellcolor{c2} 63.5 &  73.1 &  \bf \cellcolor{c8} 54.6 &  \bf \cellcolor{c14} 48.2 &  \cellcolor{c1} 60.0 \\ 
SD              & 71.5 &  \cellcolor{c6} 62.2 &  77.5 &  \bf \cellcolor{c26} 64.5 &  \bf \cellcolor{c40} 57.2 &  \bf \cellcolor{c6} 82.3 &  \bf \cellcolor{c1} 74.8 &  \cellcolor{c2} 64.0 &  \bf 75.8 &  \cellcolor{c19} 47.1 &  \cellcolor{c21} 41.1 &  \bf \cellcolor{c6} 58.2 &  \bf \cellcolor{c1} 73.7 & \bf  \cellcolor{c3} 64.2 &  \bf \cellcolor{c1} 74.9 &  \cellcolor{c6} 53.2 &  \cellcolor{c9} 43.4 &  \bf \cellcolor{c5} 63.3 \\ 
    \end{tabular}
    }
    \caption{Test results of group-robust algorithms in \textbf{\emph{head} and \emph{tail} classes} in the medium-sized datasets. \emph{Head} are the 50\% most represented (frequent) classes in the training set, and \emph{tail} are the bottom 50\%.}
    \label{tab:frequency}
    \vspace{-2mm}
\end{table*}

\paragraph{Main Results}

To highlight the temporal concept drift, we initially fine-tune BERT in all datasets with the standard ERM optimization algorithm using both \emph{random} and \emph{chronological} splits. Table~\ref{tab:data_splits} shows that the real-world performance achieved using the chronological split is severely overestimated using the random split (approx.~+10\% across evaluation measures) in two out of threee datasets. While all datasets have inherently skewed distributions (class imbalance), which is naturally demonstrated by the performance discrepancy between \microf and \macrof scores (especially when we consider the larger label sets), the temporal dimension further exacerbate the performance discrepancy as label distributions also vary across subsets (Figure~\ref{fig:imbalance}). Surprisingly, the performance discrepancy between chronological and random splits is much lower on BIOASQ (approx. 1-2\%), which could be explained by the larger volume of training data (Table~\ref{tab:datasets}), and the very high representation for most of the labels in general (Figure~\ref{fig:imbalance}).

\begin{figure*}[t]
    \centering
    \includegraphics[width=\textwidth]{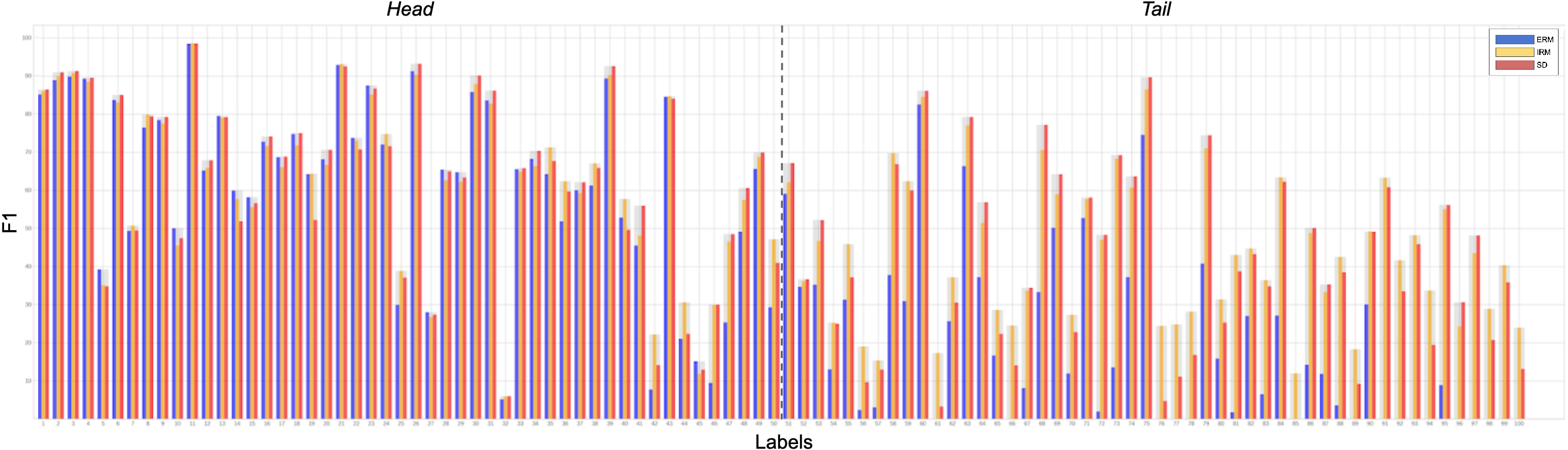}
    \vspace{-5mm}
    \caption{\textbf{\emph{Class-wise} F1-score} results for ERM (blue), IRM (yellow) and Spectral Decoupling (red) on medium-sized EUR-LEX. The classes have been ordered (left-to-right) based on the label distribution in the training subset. Algorithms' performance on the left part (\textbf{\emph{head} classes}) is very much aligned, contrary to the right (\textbf{\emph{tail} labels}).}
    \label{fig:class_f1}
    \vspace{-1mm}
\end{figure*}

In Table~\ref{tab:overall}, we present the overall results for the different optimization algorithms considering the baseline model, BERT. We observe that using a \emph{group sampler} (ERM+GS), which equals standard oversampling of minority classes, slightly improve the results in \macrof (+1-4\%) in many cases, while the performance is comparable in \microf and \mrp.  Considering the results of group-robust algorithms, we observe that most of them improve \macrof across datasets compared to ERM and ERM+GS, +1-4\% for small-sized datasets and +5-12\% in medium-sized datasets. Again the performance in \microf and \mrp is mostly comparable or a bit lower, as sample-wise averaged measures are dominated by frequent classes due to class imbalance. 

Contrary, Group DRO is consistently outperformed even by the standard ERM. Recall that Group DRO uses a weighted average of the group-wise (label-wise) losses (Equation~\ref{eq:dro_loss}-\ref{eq:dro_loss_w}), where the group weights rely on the momentum of the group-wise (label-wise) losses (Equation~\ref{eq:dro_loss_w}). In our case, this regularization acts counter-intuitively, as weights for the infrequent classes, which are rarely present across batches, are not updated (decrease) constantly. This leads to an asymmetry, where some weights are frequently updated, while others not, and in time the latter are almost zeroed-out and not affect the training objective (loss).

\paragraph{The effect of group-robust algorithms in relation to the size of the label set.}

In Table~\ref{tab:overall}, we can also observe that the performance gains of group-robust algorithms compared to ERM are greater when we use the larger label sets. This is also as the class imbalance and temporal concept drift are more severe when we consider more refined labels, especially considering \macrof. 

\paragraph{The effect of group-robust algorithms in relation to class frequency.}

In Table~\ref{tab:frequency}, we present results for the different optimization algorithms considering two groups of classes based on their frequency. \emph{Head} classes are the 50\% most frequent classes in the training set, while \emph{tail} are the bottom 50\%. As expected, the performance in head classes is much better compared to tail ones across datasets (approx.\ +20-40\% in \macrof). We observe that the performance gains of group-robust algorithms compared to ERM are greater in the tail classes (+10-40\% in \macrof). This is further highlighted in Figure~\ref{fig:class_f1}, where we observe that IRM and Spectral Decoupling, the two best performing group-robust algorithms, have larger gains in the right part (tail labels); in fact ERM scores zero in many cases (classes) where the two group-robust algorithms don't. This is highly expected as the goal of the group-robust algorithms is to minimize the group-wise (in our case, label-wise) disparity. Group DRO is severely out-performed in both head and tail, especially in the tail classes (whose weights have been zeroed-out, as previously noticed).

\begin{table*}[t]
    \centering
    \resizebox{\textwidth}{!}{
    \begin{tabular}{l|ccc|ccc|ccc|ccc|ccc|ccc}
        \multirow{2}{*}{\bf Algorithm} & \multicolumn{9}{c|}{\bf \textsc{BERT}}  & \multicolumn{9}{c}{\bf \textsc{BERT-LWAN}}  \\
        & \multicolumn{3}{c|}{\emph{Overall}} & \multicolumn{3}{c|}{\emph{Head}} & \multicolumn{3}{c|}{\emph{Tail}} & \multicolumn{3}{c|}{\emph{Overall}} & \multicolumn{3}{c|}{\emph{Head}} & \multicolumn{3}{c}{\emph{Tail}} \\
        & \microf & \macrof & \mrp & \microf & \macrof & \mrp & \microf & \macrof & \mrp & \microf & \macrof & \mrp & \microf & \macrof & \mrp & \microf & \macrof & \mrp \\
        \hline
ERM             & 68.1 &  40.7 &  71.7 &  73.4 &  61.9 &  75.7 &  27.5 &  19.4 &  51.7 & 70.5 &  49.0 &  72.3 &  74.7 &  64.3 &  75.9 &  43.0 &  33.7 &  54.0 \\ 
ERM+GS          & \cellcolor{c1} 69.3 &  \cellcolor{c13} 54.7 &  71.2 &  73.3 &  \cellcolor{c1} 63.9 &  \cellcolor{n1} 74.2 &  \cellcolor{c20} 48.1 &  \cellcolor{c26} 45.4 &  \cellcolor{c4} 56.3  & \cellcolor{n1} 68.9 & \cellcolor{c5} 53.8 & \cellcolor{n1} 71.1 & \cellcolor{n1} 73.4 & \cellcolor{n1} 63.6 & \cellcolor{n3} 73.2 & \cellcolor{c3} 45.7 & \cellcolor{c7} 41.2 & \cellcolor{c3} 57.3 \\ 
\hline
Group Uniform   & 68.9 &  \cellcolor{c9} 50.4 &  71.2 &  73.4 &  62.3 &  74.7 &  \cellcolor{c15} 42.7 &  \cellcolor{c19} 38.5 &  \cellcolor{c1} 53.0 & \cellcolor{n1} 68.7 &  \cellcolor{c5} 54.6 &  \cellcolor{n1} 70.8 &  \cellcolor{n1} 72.8 &  63.4 &  \cellcolor{n1} 74.3 &  \cellcolor{c4} 48.0 &  \cellcolor{c12} 45.8 &  \cellcolor{c2} 56.1 \\ 
Group DRO       & \cellcolor{n4} 63.4 &  \cellcolor{n12} 27.8 &  \cellcolor{n8} 63.3 &  \cellcolor{n3} 70.0 &  \cellcolor{n11} 50.6 &  \cellcolor{n5} 70.2 &  \cellcolor{n20} 7.1 &  \cellcolor{n14} 4.9 &  \cellcolor{n22} 28.8 & \cellcolor{n3} 66.8 &  \cellcolor{n9} 39.8 &  \cellcolor{n6} 65.9 &  \cellcolor{n2} 72.1 &  \cellcolor{n4} 59.4 &  \cellcolor{n5} 70.7 &  \cellcolor{n11} 31.0 &  \cellcolor{n13} 20.2 &  \cellcolor{n10} 43.6 \\ 
Deep CORAL      & 67.9 &  \cellcolor{c4} 45.2 &  \cellcolor{n1} 70.2 &  72.6 &  \cellcolor{n1} 60.1 &  \cellcolor{n2} 73.4 &  \cellcolor{c8} 35.8 &  \cellcolor{c10} 30.4 &  \cellcolor{c5} 56.8 & \multicolumn{3}{c|}{n/a}  &  \multicolumn{3}{c|}{n/a} &  \multicolumn{3}{c}{n/a} \\ 
V-REx           & 68.8 &  \cellcolor{c8} 49.2 &  \cellcolor{n2} 69.6 &  73.1 &  61.7 &  \cellcolor{n2} 73.3 &  \cellcolor{c15} 42.6 &  \cellcolor{c17} 36.8 &  \cellcolor{c3} 55.3 & \cellcolor{n1} 69.2 &  \cellcolor{c6} 55.0 &  \cellcolor{n2} 70.1 &  \cellcolor{n1} 73.1 &  63.9 &  \cellcolor{n1} 74.2 &  \bf \cellcolor{c5} 48.7 &  \bf \cellcolor{c12} 46.1 &  \bf \cellcolor{c4} 58.4 \\ 
IRM             & \cellcolor{c2} 70.4 &  \bf \cellcolor{c14} 54.8 &  72.4 &  74.4 &  \cellcolor{c2} 64.4 &  75.2 &  \cellcolor{c21} 48.7 &  \cellcolor{c25} 45.1 &  \cellcolor{c4} 56.5 & \cellcolor{n1} 69.1 &  \cellcolor{c3} 53.0 &  71.6 &  \cellcolor{n1} 73.2 &  \cellcolor{n1} 63.2 &  \cellcolor{n1} 74.8 & \cellcolor{c4} 47.0 &  \cellcolor{c9} 42.8 &  \cellcolor{c2} 56.5 \\ 
SD              & \cellcolor{c2} 70.8 &  \cellcolor{c11} 52.5 &  \bf \cellcolor{c1} 72.7 & \cellcolor{c1} 74.8 &  \cellcolor{c2} 64.0 &  \bf 75.8 &  \cellcolor{c19} 47.1 &  \cellcolor{c21} 41.1 &  \cellcolor{c6} 58.2 & 70.4 &  \cellcolor{c5} 54.5 &  \cellcolor{n1} 70.4 &  74.4 &  64.6 &  \cellcolor{n2} 73.3 &  \cellcolor{c4} 47.8 &  \cellcolor{c10} 44.5 &  \cellcolor{c4} 58.5 \\ 
\hline
LW-DRO (v1)     & \cellcolor{c1} 69.9 &  \cellcolor{c5} 46.3 &  \cellcolor{n3} 68.4 &  \cellcolor{c1} 74.7 &  62.6 &  \cellcolor{n1} 73.8 &  \cellcolor{c11} 39.4 &  \cellcolor{c10} 30.1 &  \cellcolor{n6} 45.5 & \cellcolor{n1} 69.8 & \cellcolor{c4} 53.4 & \cellcolor{n3} 69.3  & \cellcolor{n1} 74.1 & \cellcolor{n1} 63.2 & \cellcolor{n4} 71.5 & \cellcolor{n2} 41.4 & \cellcolor{c6} 39.7 & \cellcolor{n2} 52.0 \\ 
LW-DRO (v2)     &  \cellcolor{c3} 71.3 &  \cellcolor{c13} 54.2 &  \cellcolor{n1} 70.3 &  \bf \cellcolor{c1} 75.3 &  \cellcolor{c3} 65.1 &  \cellcolor{n1} 74.0 &  \bf \cellcolor{c21} 49.2 &  \cellcolor{c23} 43.3 &  \cellcolor{c1} 53.4 
& \bf 71.5 &  \cellcolor{c5} 54.0 &  \cellcolor{n1} 70.5 &  \cellcolor{n1} 74.1 &  \bf \cellcolor{c1} 65.5 &  \cellcolor{n1} 74.0 &  \cellcolor{c5} 48.4 &  \cellcolor{c10} 43.9 & \cellcolor{c3}  56.6 \\  

    \end{tabular}
    }
    \caption{Test results of group-robust algorithms with \textbf{different models} (BERT, and BERT-LWAN) in the medium-sized version of \textsc{EUR-LEX}. Deep CORAL is not applicable (n/a) in LWAN -there is not a universal featurizer-.}
    \label{tab:models}
    \vspace{-4mm}
\end{table*}

\paragraph{Why IRM and Spectral Decoupling are a better fit compared to the rest of the algorithms?}

To answer this question, we need to identify the main differentiation between IRM, Spectral Decoupling and the rest of the methods. Both IRM and Spectral Decoupling follow similar incentives. IRM penalizes variance across losses in the same group (Equation~\ref{eq:irm_loss}), i.e., in our case, the network is penalized if there is a performance disparity between samples labeled with the same classes using as a reference a dummy classifier. Spectral Decoupling penalizes the variance across label predictions (Equation~\ref{eq:sd_loss}), i.e., the network is penalized for being over-confident. The rest of the algorithms mainly rely on an equal consideration of the group-wise (in our case, label-wise) losses (Equation~\ref{eq:mean_loss}), i.e., in our case, all classes are equally important for the training objective. 

The latter incentive (averaging across group-wise losses) seems very intuitive, although in practice the groups (labels) co-occur (are not mutually exclusive) in a multi-label setting, thus frequent labels remain ``first class citizens'' in the optimization process, biasing parameter updates in their favor.  

Contrary, both IRM and Spectral Decoupling use a learning component (loss term), which penalizes \emph{label degeneration}. This is particularly important in multi-label classification, especially when we consider large label sets, as networks tend to over-fit (specialize) in few dominant (frequent) labels that shape the training loss and finally ignore (zero-out) the rest of the labels. This is quite different from the concept of \emph{Gradient Starvation}, introduced by \citet{pezeshki2020gradient}, where a network becomes over-confident in its predictions by capturing only few dominant features, as in our case the main issue is the label degeneration rather than possible spurious correlations learned by the network. Moreover, Spectral Decoupling does not rely on group-wise losses, similar to the rest.

\paragraph{The effect of group-robust algorithms using BERT-LWAN.}

In this part, we compare the effect of the group-robust algorithms in between standard BERT and BERT-LWAN on the medium-sized EUR-LEX dataset. In Table~\ref{tab:models}, we observe that BERT-LWAN closes the gap between ERM and the best-of group-robust algorithms. The results of ERM when we use BERT-LWAN are improved across measures, especially when we consider \macrof with a 10\% improvement over the standard BERT. Both IRM and Spectral Decoupling seem quite insensitive to the underlying model. Similarly, the results for the rest of the group-robust algorithms are improved. Nonetheless, there are still benefits in \macrof and less represented (\emph{tail}) labels in general. Interestingly, Spectral Decoupling improves results in \macrof, with comparable \microf scores. Although, we observe a performance drop (approx. 2\%) in \mrp when we consider overall and head classes. We hypothesize that IRM and Spectral Decoupling negatively affect the ability of the BERT-LWAN model to correctly rank labels (Equation~\ref{eq:lwan_outputs}), as they force the model to consider all labels by not being over-confident (discriminatory) with one way or another, as previously explained.

\begin{figure}[t]
    \centering
    \includegraphics[width=\columnwidth]{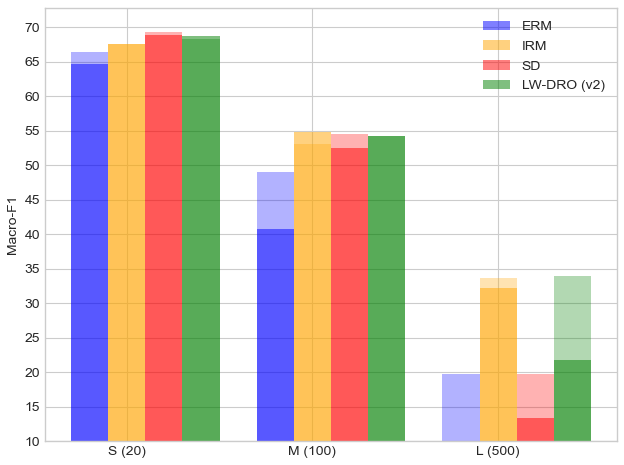}
    \caption{LWAN-BERT test performance using ERM, IRM, Spectral Decoupling (SD), and LW-DRO (v2) across \textbf{all EUR-LEX settings}. The shaded part of the bars denotes the performance improvement (of LWAN-BERT) compared to the standard BERT.}
    \label{fig:eurlex_lwan}
    \vspace{-2mm}
\end{figure}

In Figure~\ref{fig:eurlex_lwan}, we compare the performance of ERM, IRM, and Spectral Decoupling across three EUR-LEX settings, small-sized, medium-sized, and one extra large-sized considering the 3rd level of EuroVoc including 500 concepts (labels). In the small label set, we observe that the use of LWAN-BERT slightly improves the performance when trained with ERM compared to standard BERT (shaded part of the bars). In the medium label set, as already discussed, we observe an approx.\ 10\% improvement with ERM, while in case of the large label set, using LWAN-BERT leads to an approx.\ 20\% improvement with ERM (the performance of BERT is 0\%), and 6.5\% with Spectral Decoupling, while IRM proves to be remarkably robust across all settings and both neural methods (BERT with or without the LWAN component).

\section{Alternative Combined Algorithm}

Having a clear understanding of what IRM and Spectral Decoupling offer, it seems that we could combine both to leverage all features: (a) rely on group-wise (label-wise) losses as the main driver of the optimization process (Equation~\ref{eq:mean_loss}); (b) penalize the classifier if there is a performance disparity between samples labeled with the same classes (Equation~\ref{eq:irm_loss}); and (c) penalize the classifier for being over-confident (Equation~\ref{eq:sd_loss}). 

We name the new algorithm \emph{Label-Wise Distributional Robust Optimization}, LW-DRO in short, as it mainly aims to mitigate label-wise disparities, and investigate two alternatives (variants):

\begin{itemize}[leftmargin=8pt]
    \item In version 1 (\emph{v1}), we combine the averaged group-wise (label-wise) losses (Equation~\ref{eq:mean_loss}) introduced with Group Uniform, with the Spectral Decoupling penalty (Equation~\ref{eq:sd_loss}). The total loss term ($\mathcal{L}_{LW-DRO}$), is computed as follows:
    \begin{equation}
    \frac{1}{G} \sum_{i=1}^{G} \mathcal{L}(g_i) + \lambda * \frac{1}{N}\sum_{i=1}^{N} {\hat{y}_i}^2
    \label{eq:lwdro1}
\end{equation}
    \item In version 2 (\emph{v2}), we also include the group-wise penalties of IRM (Equation~\ref{eq:irm_loss}). The total loss term ($\mathcal{L}_{LW-DRO}$), is computed as follows:
\begin{equation}
    \frac{1}{G} \left(\sum_{i=1}^{G} \mathcal{L}(g_i) + \lambda_1 * P(g_i)\right) + \lambda_2 * \frac{1}{N}\sum_{i=1}^{N} {\hat{y}_i}^2
    \label{eq:lwdro2}
\end{equation}
\end{itemize}

\noindent The notation used in Equations~\ref{eq:lwdro1} and \ref{eq:lwdro2} follows the one presented in Section~\ref{sec:algorithms}.\vspace{1mm}

In Table~\ref{tab:models}, we observe that the second variant of LW-DRO (v2) has comparable or better performance compared to IRM and Spectral Decoupling, contrary to the first one (v1). LW-DRO (v2) is a straight forward combination of IRM and  Spectral Decoupling, while LW-DRO (v1) that relies on a group-averaged loss under-performs, especially considering the \macrof scores. As previously explained, labels co-occur in a multi-label setting, hence averaging label-wise losses favors frequent classes and in turn limits the possible benefits in under-represented classes (perceived by \macrof).

In Figure~\ref{fig:eurlex_lwan}, we present the results of ERM, and the 3 overall best group-robust algorithms (IRM, Spectral Decoupling, and LW-DRO (v2)) across all EUR-LEX settings. LW-DRO (v2) has comparable performance in the first two setting (small, medium), while being slightly better than IRM in the large-sized setting. While LW-DRO (v2) seems to control the trade-offs between IRM and Spectral Decoupling, we believe that future work should better seek alternative directions with respect to algorithmic advances that possibly mitigate label degeneration and tackles label-wise disparities.

\section{Conclusions \& Future Work}

We considered one of the main challenges in large-scale multi-label text classification, which comes from the fact that not all labels are well represented in the training set due to the class imbalance and the effect of temporal concept drift.  To mitigate label disparities, we considered several group-robust optimization algorithms initially proposed to mitigate group disparities given specific attributes. 
Experimenting with three datasets in two different settings, we empirically find that group-robust algorithms vastly improve performance considering macro-averaged measures, while two of the group-robust algorithms (Invariant Risk Minimization and Spectral Decoupling) improve performance across all measures. Considering a more well-suited neural method (LWAN-BERT), we observe a vast performance improvement using ERM, leading to comparable overall results (\microf, \mrp) with the group-robust algorithms; although is still outperformed considering \macrof. 
Lastly, based on our understanding of what IRM and Spectral Decoupling, the two best group-robust algorithms, offer, we introduced and evaluated a new algorithm, Label-Wise DRO, which combines features from both, and one of its variants has comparable or better performance considering larger label sets.

In the future, we would like to further investigate the two-tier anomaly (class imbalance and temporal concept drift). In this direction, we would like to directly take into consideration the time dimension by utilizing this information in group sampling and algorithms (e.g., groups over period of time). 
We would also like to consider data augmentation techniques (e.g., paraphrasing via masked-language modeling \cite{ng-etal-2020-ssmba}, and teacher forcing exploiting unlabeled data \cite{eisenschlos-etal-2019-multifit}) to improve the data (feature) sampling variability, as the group sampler used in group-robust algorithms over-sample minority classes with the same limited instances. 
Further on, we would like to investigate the use of zero-shot LWAN methods \cite{rios-kavuluru-2018-shot, chalkidis-etal-2020-empirical}, which currently harm averaged performance in favor of improved worst case performance. Label encodings based on contextualized word representations generated by pre-trained language models \cite{hardalov-etal-2021} may mitigate the effect of using non-contextualized ones (e.g., Word2Vec).

\section*{Acknowledgments} This work is funded by the Innovation Fund Denmark (IFD)\footnote{\url{https://innovationsfonden.dk/en}} under File No.\ 0175-00011A.

\bibliography{anthology, custom}
\bibliographystyle{acl_natbib}

\appendix

\begin{figure*}[t]
    \centering
    \includegraphics[width=0.9\textwidth]{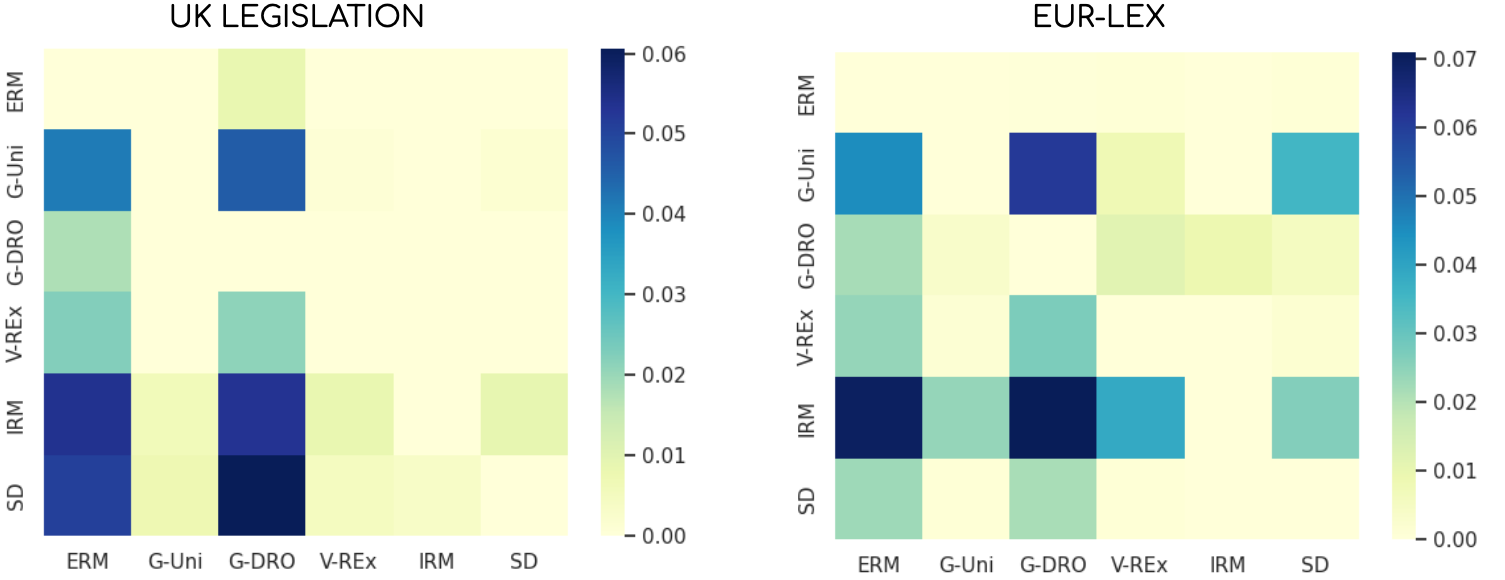}
    \caption{\textbf{\emph{Class-wise bias}} in-between algorithms across datasets, measured with the normalized Combined Error Variance (CEV) as defined by \citet{blakeney2021measure}.}
    \label{fig:cev}
    \vspace{-3mm}
\end{figure*}

\section{Measuring class-wise bias}

\citet{blakeney2021measure} recently introduced two evaluation measures to estimate class-wise bias of two models in comparison to one another in a multi-class setting, and show that these metrics can be also used to measure fairness and bias with respect to protected attributes.

Following \citet{blakeney2021measure}, in Figure~\ref{fig:cev} we present the normalized Combined Error Variance (CEV) in-between algorithms. CEV estimates the class-wise bias of a model A relative to another model B has increased of the change between model A and a random predictor. For a detailed analysis of the CEV metric, please refer to \citet{blakeney2021measure}. 

In our case, as different models, we consider BERT trained with a different algorithm. In both UK-LEX and EUR-LEX,  swapping Group Uniform, IRM, or Spectral Decoupling with ERM, or Group DRO leads to a higher class-wise bias, which is highly expected given the aforementioned performance analysis, i.e., improved \macrof scores.

\section{Additional Results}
\label{sec:add_results}

\begin{table}[h]
    \centering
    \resizebox{\columnwidth}{!}{
    \begin{tabular}{l|cc|cc}
        \multirow{2}{*}{\bf Algorithm} &  \multicolumn{2}{c|}{ \emph{Small}} & \multicolumn{2}{c}{ \emph{Medium}} \\
        & \macrof & \microf & \macrof & \microf \\
        \hline
         ERM  & 71.8 & 60.2  & 47.4 & 10.3 \\
         ERM+GS & 71.7 & \cellcolor{c2} 62.4  & 47.5 & \cellcolor{c2} 12.6 \\ 
         \hline
         Group Uniform & 71.9 & \cellcolor{c6} 66.1  & 48.2 & \cellcolor{c4} 13.3 \\
         Group DRO & \cellcolor{n6} 65.2 & \cellcolor{n10} 47.4  & \cellcolor{n20} 14.0 & \cellcolor{n6} 3.8  \\
         Deep CORAL & 72.1 & \cellcolor{c6} 67.1  & 47.1 & \cellcolor{c2} 12.3  \\
         V-REx & 71.9 & \cellcolor{c6} 65.9 & 47.6 & 11.3 \\
         IRM & 72.0 & \cellcolor{c6} 66.6  & \cellcolor{c6} \bf 53.3 & \cellcolor{c8} \bf 18.3 \\
         Spectral Decoupling & \bf 72.3 & \cellcolor{c8} \bf 67.2  & \cellcolor{c6} 53.1 & \cellcolor{c6} 16.1 \\

    \end{tabular}
    }
    \caption{\textbf{Overall} test results  of the \textbf{group-robust algorithms} across on MIMIC-III dataset.}
    \label{tab:mimic}
\end{table}

\subsection{Experiments on MIMIC-III}
\label{sec:mimic}

\paragraph{MIMIC-III} dataset \cite{Johnson2017} contains approx.\ 50k discharge summaries from US hospitals. Each summary is annotated with one or more codes (labels) from the ICD-9 hierarchy, which has 8 levels.\footnote{\url{www.who.int/classifications/icd/en/}}. The International Classification of Diseases, Ninth Revision (ICD-9)  is the official system of assigning codes to diagnoses and procedures associated with hospital utilization in the United States and is maintained by the World Health Organization (WHO). 

MIMIC-III has been anonymized to protect patients privacy, including chronological information (e.g., entry/discharge dates). Hence, it is not possible to split data in chronological splits. We split the dataset randomly in training (30k), development (10k), test (10k) subsets. We use the 1st and 2nd level of ICD-9 including 19 and 184 categories, respectively. 

In Table~\ref{tab:mimic}, we present the results, which lead to the very same observations discussed for the rest of the datasets.

\subsection{Development Results}
\label{sec:dev_results}

We run three repetitions with different random seeds and in the main article (Section~\ref{sec:results}, we report the test scores based on the seed with the best scores on development data. For completeness, in Tables~\ref{tab:dev_results_1}, ~\ref{tab:dev_results_2}, ~\ref{tab:dev_results_3}, we report the development results  of the group-robust (label-robust) algorithms across all datasets (UK-LEX, EUR-LEX, BIOASQ) and settings (small and medium sized label sets) using BERT.  We report the mean and standard deviation ($\pm$) across all three examined seeds.

\begin{table*}[t]
    \centering
    \resizebox{\textwidth}{!}{
    \begin{tabular}{l|l|l|l|l|l|l}
        \multirow{3}{*}{\bf Algorithm} & \multicolumn{2}{c|}{\bf \textsc{UK-LEX}} & \multicolumn{2}{c|}{\bf \textsc{EUR-LEX}}  & \multicolumn{2}{c}{\bf \textsc{BIO-ASQ}}\\
         & \emph{Small} & \emph{Medium}  & \emph{Small} & \emph{Medium}  & \emph{Small} & \emph{Medium} \\
         ERM             & 83.9 $\pm$ 0.4 &  72.7 $\pm$ 0.3 &  82.7 $\pm$ 0.1 &  73.9 $\pm$ 0.2 &  86.3 $\pm$ 0.1 &  68.5 $\pm$ 0.2 \\  
ERM+GS          & 83.7 $\pm$ 0.2 &  76.1 $\pm$ 0.1 &  0.0 $\pm$ 0.0 &  75.3 $\pm$ 0.1 &  85.8 $\pm$ 0.2 &  68.0 $\pm$ 0.1 \\
\hline
Group Uniform   & 83.4 $\pm$ 0.4 &  75.9 $\pm$ 0.1 &  82.5 $\pm$ 0.1 &  75.1 $\pm$ 0.3 &  85.7 $\pm$ 0.1 &  68.5 $\pm$ 0.5 \\  
Group DRO       & 83.4 $\pm$ 0.2 &  67.9 $\pm$ 0.2 &  81.6 $\pm$ 0.3 &  69.7 $\pm$ 0.2 &  84.6 $\pm$ 0.1 &  43.5 $\pm$ 6.5 \\  
Deep CORAL      & 83.2 $\pm$ 0.5 &  73.5 $\pm$ 0.2 &  82.7 $\pm$ 0.0 &  73.8 $\pm$ 0.3 &  85.3 $\pm$ 0.1 &  67.5 $\pm$ 0.3 \\  
V-REx             & 83.8 $\pm$ 0.2 &  73.9 $\pm$ 0.2 &  82.4 $\pm$ 0.1 &  75.2 $\pm$ 0.0 &  85.7 $\pm$ 0.0 &  68.2 $\pm$ 0.5 \\  
IRM             & 83.7 $\pm$ 0.6 &  77.3 $\pm$ 0.3 &  82.3 $\pm$ 0.2 &  76.0 $\pm$ 0.4 &  85.6 $\pm$ 0.1 &  69.5 $\pm$ 0.7 \\  
SD              & 84.1 $\pm$ 0.5 &  77.4 $\pm$ 0.2 &  83.1 $\pm$ 0.1 &  76.5 $\pm$ 0.2 &  86.0 $\pm$ 0.0 &  70.8 $\pm$ 0.1 \\  
\end{tabular}
}
\caption{\textbf{Overall} \microf development results  of the \textbf{group-robust (label-robust) algorithms} across all datasets (UK-LEX, EUR-LEX, BIOASQ) and settings (small and medium sized label sets). We report the mean and standard deviation ($\pm$) across three seeds.}
\label{tab:dev_results_1}
\end{table*}

\begin{table*}[t]
    \centering
    \resizebox{\textwidth}{!}{
    \begin{tabular}{l|l|l|l|l|l|l}
        \multirow{3}{*}{\bf Algorithm} & \multicolumn{2}{c|}{\bf \textsc{UK-LEX}} & \multicolumn{2}{c|}{\bf \textsc{EUR-LEX}}  & \multicolumn{2}{c}{\bf \textsc{BIO-ASQ}}\\
         & \emph{Small} & \emph{Medium}  & \emph{Small} & \emph{Medium}  & \emph{Small} & \emph{Medium} \\
ERM             & 78.9 $\pm$ 0.7 &  27.7 $\pm$ 19.6 &  67.1 $\pm$ 0.7 &  44.4 $\pm$ 0.7 &  75.8 $\pm$ 0.6 &  47.6 $\pm$ 0.6 \\  
ERM+GS          & 80.0 $\pm$ 0.4 &  47.4 $\pm$ 0.5 &  68.3 $\pm$ 0.0 &  60.4 $\pm$ 0.7 &  76.2 $\pm$ 0.2 &  49.9 $\pm$ 0.1 \\
\hline
Group Uniform   & 80.2 $\pm$ 0.4 &  66.6 $\pm$ 0.3 &  71.9 $\pm$ 0.5 &  56.6 $\pm$ 0.4 &  76.6 $\pm$ 0.1 &  52.3 $\pm$ 1.4 \\  
Group DRO       & 79.5 $\pm$ 0.3 &  35.2 $\pm$ 1.1 &  65.4 $\pm$ 2.3 &  32.2 $\pm$ 0.8 &  73.3 $\pm$ 0.6 &  13.9 $\pm$ 4.2 \\  
Deep CORAL      & 79.6 $\pm$ 0.6 &  54.3 $\pm$ 1.7 &  72.1 $\pm$ 0.0 &  49.5 $\pm$ 1.3 &  75.5 $\pm$ 0.4 &  55.7 $\pm$ 1.8 \\  
V-REx             & 80.2 $\pm$ 0.7 &  61.0 $\pm$ 0.9 &  72.0 $\pm$ 0.3 &  55.7 $\pm$ 0.2 &  76.6 $\pm$ 0.1 &  49.9 $\pm$ 1.7 \\  
IRM             & 80.2 $\pm$ 0.4 &  69.6 $\pm$ 0.7 &  71.4 $\pm$ 0.5 &  60.8 $\pm$ 1.6 &  76.7 $\pm$ 0.1 &  55.7 $\pm$ 2.2 \\  
SD              & 81.2 $\pm$ 0.8 &  69.3 $\pm$ 0.5 &  73.4 $\pm$ 0.3 &  58.8 $\pm$ 0.3 &  77.0 $\pm$ 0.2 &  54.5 $\pm$ 0.5 \\  
\end{tabular}
}
\caption{\textbf{Overall} \macrof development results  of the \textbf{group-robust (label-robust) algorithms} across all datasets (UK-LEX, EUR-LEX, BIOASQ) and settings (small and medium sized label sets).  We report the mean and standard deviation ($\pm$) across three seeds.}
\label{tab:dev_results_2}
\end{table*}

\begin{table*}[t]
    \centering
    \resizebox{\textwidth}{!}{
    \begin{tabular}{l|l|l|l|l|l|l}
        \multirow{3}{*}{\bf Algorithm} & \multicolumn{2}{c|}{\bf \textsc{UK-LEX}} & \multicolumn{2}{c|}{\bf \textsc{EUR-LEX}}  & \multicolumn{2}{c}{\bf \textsc{BIO-ASQ}}\\
         & \emph{Small} & \emph{Medium}  & \emph{Small} & \emph{Medium}  & \emph{Small} & \emph{Medium} \\
ERM             & 87.5 $\pm$ 0.5 &  77.2 $\pm$ 0.6 &  86.1 $\pm$ 0.0 &  75.5 $\pm$ 0.8 &  88.3 $\pm$ 0.0 &  71.0 $\pm$ 0.3 \\  
ERM+GS          & 88.7 $\pm$ 0.3 &  77.6 $\pm$ 0.4 &  86.5 $\pm$ 0.1 &  75.8 $\pm$ 0.6 &  89.4 $\pm$ 0.2 &  70.5 $\pm$ 0.1 \\  
\hline
Group Uniform   & 87.0 $\pm$ 0.4 &  80.1 $\pm$ 0.3 &  85.2 $\pm$ 0.6 &  75.7 $\pm$ 0.7 &  87.4 $\pm$ 0.1 &  70.1 $\pm$ 0.4 \\  
Group DRO       & 86.6 $\pm$ 0.3 &  75.0 $\pm$ 0.2 &  82.6 $\pm$ 0.4 &  69.7 $\pm$ 0.2 &  85.7 $\pm$ 0.2 &  43.1 $\pm$ 6.9 \\  
Deep CORAL      & 87.0 $\pm$ 0.3 &  78.3 $\pm$ 0.7 &  85.7 $\pm$ 0.1 &  75.7 $\pm$ 0.2 &  86.1 $\pm$ 0.3 &  70.9 $\pm$ 0.7 \\  
V-REx             & 87.5 $\pm$ 0.2 &  79.9 $\pm$ 0.6 &  85.5 $\pm$ 0.3 &  75.6 $\pm$ 0.0 &  87.4 $\pm$ 0.0 &  70.0 $\pm$ 0.4 \\  
IRM             & 87.2 $\pm$ 0.4 &  80.7 $\pm$ 0.3 &  85.3 $\pm$ 0.4 &  76.4 $\pm$ 0.8 &  87.4 $\pm$ 0.0 &  70.5 $\pm$ 0.4 \\  
SD              & 87.5 $\pm$ 0.1 &  81.1 $\pm$ 0.2 &  83.9 $\pm$ 0.2 &  76.8 $\pm$ 0.1 &  87.5 $\pm$ 0.0 &  72.6 $\pm$ 0.2 \\  
\end{tabular}
}
\caption{\textbf{Overall} \mrp development results  of the \textbf{group-robust (label-robust) algorithms} across all datasets (UK-LEX, EUR-LEX, BIOASQ) and settings (small and medium sized label sets).  We report the mean and standard deviation ($\pm$) across three seeds.}
\label{tab:dev_results_3}
\end{table*}

\end{document}